\documentclass[10pt,twocolumn,letterpaper]{article}

\usepackage{iccv}
\usepackage{times}
\usepackage{epsfig}
\usepackage{graphicx}
\usepackage{amsmath}
\usepackage{amssymb}
\usepackage{subfigure}
\usepackage{color}
\usepackage{fontawesome}
\usepackage{multirow}
\usepackage{longtable}
\usepackage{array}
\usepackage{makecell}
\usepackage{threeparttable}
\usepackage{booktabs}
\usepackage{colortbl}
\usepackage{siunitx}
\usepackage{mhchem}

\usepackage{bbding}

\usepackage[pagebackref=true,breaklinks=true,letterpaper=true,colorlinks,bookmarks=false]{hyperref}

\iccvfinalcopy % *** Uncomment this line for the final submission

\ificcvfinal\fi

\begin{document}

\title{Parsing Table Structures in the Wild}

\author{
Rujiao Long\footnotemark[1]$~^3$,\quad Wen Wang\footnotemark[1]$~^{1,2}$, \quad Nan Xue$^1$ \quad Feiyu Gao$^3$, \\ Zhibo Yang$^3$, \quad Yongpan Wang$^3$, \quad Gui-Song Xia\footnotemark[2]$~^{1,2}$\\
$^1$ {\em School of Computer Science, Wuhan University, Wuhan, China} \\
$^2$ {\em LIESMARS, Wuhan University, Wuhan, China} \\
$^3$ {\em Alibaba-Group, Hangzhou, China}\\
\faicon{github}~\url{https://github.com/wangwen-whu/WTW-Dataset}\vspace{2pt}\\
}

\maketitle
\renewcommand{\thefootnote}{\fnsymbol{footnote}}
\footnotetext[1]{Equal Contribution.} 
\footnotetext[2]{Correspondence Author.}

%%%%%%%%% ABSTRACT
\begin{abstract}
This paper tackles the problem of table structure parsing (TSP) from images in the wild. In contrast to existing studies that mainly focus on parsing well-aligned tabular images with simple layouts from scanned PDF documents, we aim to establish a practical table structure parsing system for real-world scenarios where tabular input images are taken or scanned with severe deformation, bending or occlusions. For designing such a system, we propose an approach named Cycle-CenterNet on the top of CenterNet with a novel cycle-pairing module to simultaneously detect and group tabular cells into structured tables. In the cycle-pairing module, a new pairing loss function is proposed for the network training. Alongside with our Cycle-CenterNet, we also present a large-scale dataset, named Wired Table in the Wild (WTW), which includes well-annotated structure parsing of multiple style tables in several scenes like photo, scanning files, web pages, \emph{etc.}. In experiments, we demonstrate that our Cycle-CenterNet consistently achieves the best accuracy of table structure parsing on the new WTW dataset by 24.6\% absolute improvement evaluated by the TEDS metric. A more comprehensive experimental analysis also validates the advantages of our proposed methods for the TSP task. 

\end{abstract}

%%%%%%%%% BODY TEXT
\section{Introduction}
% We should clarify the relationship of table structure parsing, table detection and table structure recognition.
% Table detection 

% \textcolor{red}{
% Tabular images are everywhere in our life to summarize and display important data into columns and rows. 
%The previous works only focus on the document images. Instead, we are interested in 
Tables are commonly used in our daily life to record and summarize important data for quick and better visualization of information. 
With the increasing popularity of smartphones and portable cameras, it is very common to share information with photo of tables. Accordingly, it is highly demanded to automatically extract and parse table structures from photos or images in the wild. 
%In this paper, we aim at parsing table structures in such a challenging configuration.
\begin{figure}
\centering
\resizebox{1\linewidth}{!}{
\begin{tabular}{p{1pt}cc}
    \raisebox{30pt}{\rotatebox[origin=c]{90}{Input}}& \includegraphics[height=0.11\textheight]{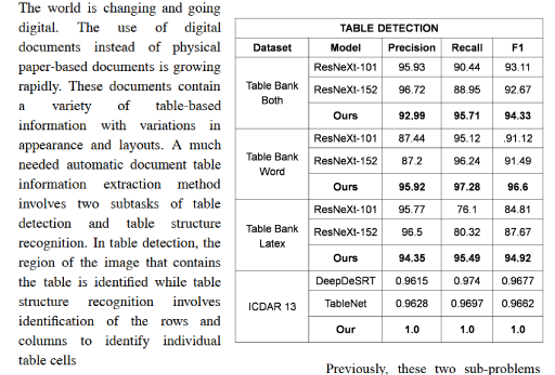} & \includegraphics[height=0.11\textheight]{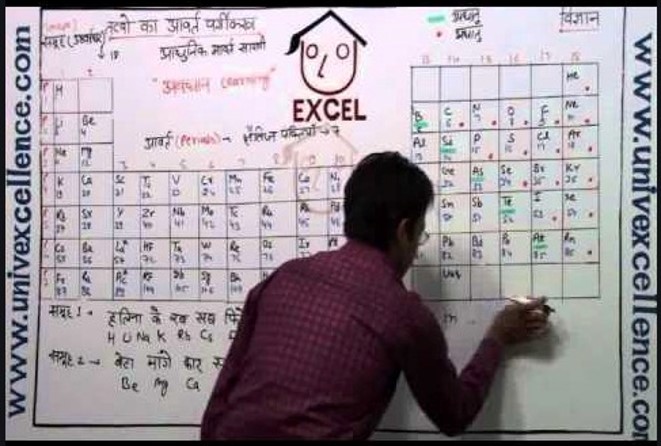} \\
    \raisebox{30pt}{\rotatebox[origin=c]{90}{TabSplit~\cite{tensmeyer2019deep}}}& \includegraphics[height=0.11\textheight]{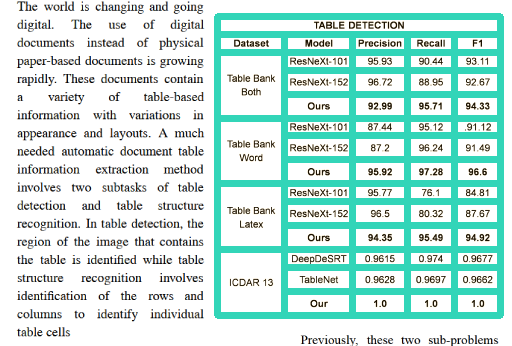} &
    \includegraphics[height=0.11\textheight]{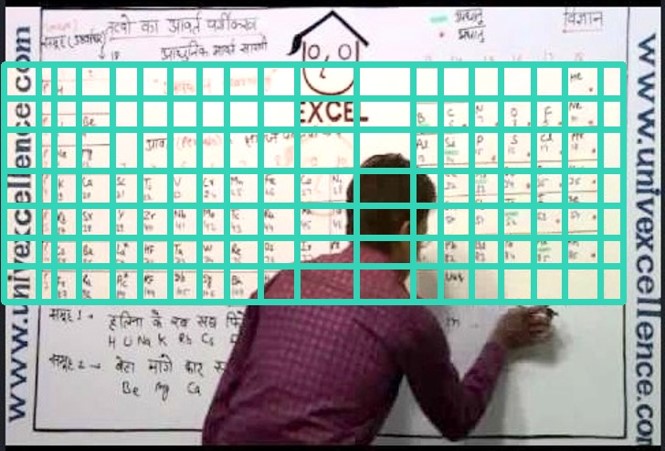}\\%\vspace{2mm}
    \raisebox{30pt}{\rotatebox[origin=c]{90}{Our Results}}&
    \includegraphics[height=0.11\textheight]{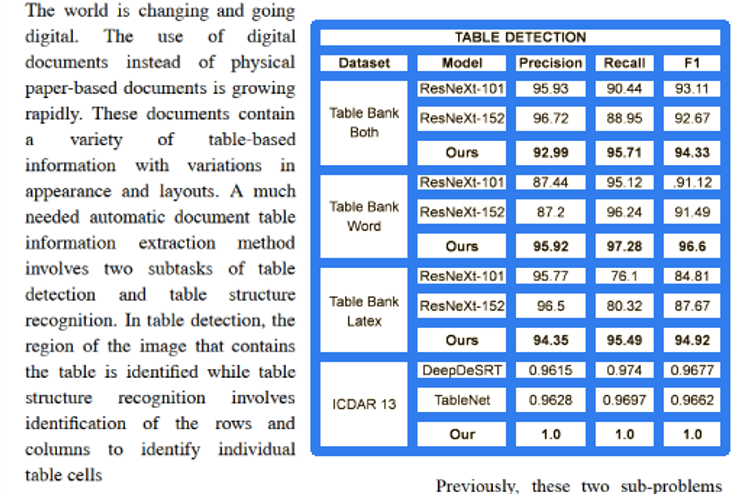} &
    \includegraphics[height=0.11\textheight]{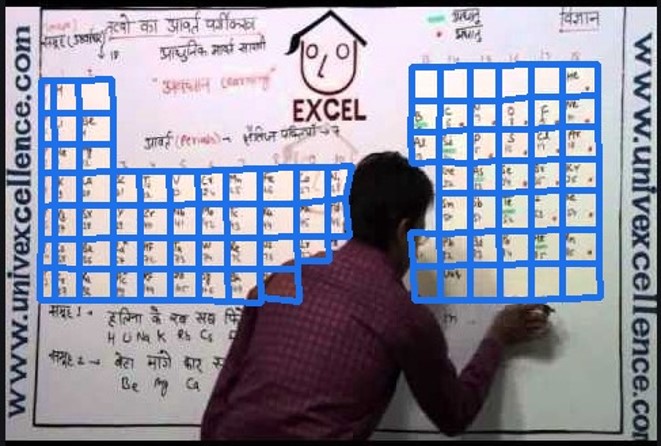}\\
    & TSP in Document images & TSP in the wild
\end{tabular}
}
\vspace{1mm}
% \subfigure{\includegraphics[width=8cm]{figs/fig1new.jpg}}
% \caption{Cycle-Centernet: A real-scene based table structure recognition method}
\caption{Visual comparison for the difference between the problem of table structure parsing (TSP) in document images and the images taken in the wild. We leverage the state-of-the-art approach for document images proposed in~\cite{tensmeyer2019deep} and our proposed Cycle-CenterNet for both input images to obtain the parsing results.}
\vspace{2mm}
\label{fig:1}
\end{figure}

Given an image, {\em Table Structure Parsing} (TSP) aims at extracting all the tables, locating their cells, and obtaining the row-column information in the image.
Previously, this problem is studied as table structure recognition focusing on document images. 
In such scenario, the tabular images are taken with well imaging conditions and are often horizontally (or vertically) aligned with clean background and clear table structures. 
Early pioneering works, \eg~\cite{itonori1993table,kieninger1998table,tupaj1996extracting,green1995recognition}, tackle the TSP problem in a bottom-up manner by heuristically grouping detected cells based on low-level cues (\eg, lines, boundaries and word regions). Recently, deep learning-based approaches are presented to avoid the heuristic grouping scheme design and resort to developing end-to-end models. However, limited by the training datasets ~\cite{li2019tablebank,shahab2010open,chi2019complicated,zhong2019image,gobel2013icdar} used for table structure parsing, they still addressed this problem under the well-aligned assumption of tabular images. 

For a more practical requirement of parsing table structures from images taken by hand-held cameras in the wild, the existing state-of-the-art approaches ~\cite{qasim2019rethinking,raja2020table,paliwal2019tablenet,schreiber2017deepdesrt,li2019tablebank,zheng2021global} are prone to fail as the commonly-used assumption of tabular images no longer holds. Specifically, the tabular images in the widely-used datasets (\eg, ICDAR-2013~\cite{gobel2013icdar}, Tablebank~\cite{li2019tablebank}) are usually with clean background and clear table structures. 
Limited by this, existing TSP approaches can only handle table structure parsing in a relative simple scenario by grouping detected cells into tables~\cite{paliwal2019tablenet,schreiber2017deepdesrt,li2019tablebank,zheng2021global}. Moreover, few research pays attention to the precision of cell boundary, which is important in text recognition. 
% \definecolor{mypink}{rgb}{.925,.90,.999}

To tackle the TSP problem in the wild, we present a large-scale dataset in this paper to address the data lacking issue. When we collecting the real-scene tabular images, we found that the wired tables and wireless tables have a very large difference. The wireless tables in natural images are very challenging to be recognized as the lacking of reference for perceptual grouping by human annotators. Therefore, we mainly focus on the challenging wired tables for annotation. 
Our proposed dataset, the {\em Wired Tables in the Wild} (WTW), contains 14,581 images with the annotated information of {\em table id}, {\em tabular cells} and {\em corresponding row/column information}. Following the data splitting strategy used in ICDAR 2019~\cite{gao2019icdar}, we split our WTW into \emph{training}/\emph{testing} subsets with 10,970 and 3611 data samples respectively.

As shown in Fig.~\ref{fig:1}, the images in the WTW dataset are very different from the document images, which thus poses a new problem to the table structures parsing task. For instance, the non-rigid image deformation and complicated image background presented in natural images will challenge existing approaches~\cite{raja2020table} for document images on detecting and grouping the tabular cells. 

With our proposed WTW dataset available, we address the problem of table structure parsing in the wild by proposing a simple yet effective approach Cycle-CenterNet.
It simultaneously detects the vertices and center points of tabular cells, and groups the cells into tables by learning the common vertices. Specifically, we found that the center point and vertices of a cell have a mutual-directed relationship that can be used to group the cells into tables by using the common vertex that is located in the intersect of the adjacent cells. Based on this, we propose a loss function named \emph{pairing loss} to end-to-end group the cells in training phase. Once the structures of tables are obtained, we use a simple post-processing algorithm to retrieve the row and column information for the parsed tables. 
% To the best of our knowledge, 
% Cycle-Centernet is the first way to process real-scene tables and achieve the state of the art.
In experiments, we evaluate the proposed Cycle-CenterNet on WTW dataset. Compared with the strong baseline of vanilla CenterNet, our approach largely improves the F1-score of physical coordinate accuracy from $73.1\%$ to $78.3\%$, while improves the F1-score of adjacent relationship estimation from $84.8\%$ to $92.4\%$. In the metric of TEDS~\cite{zhong2019image}, the proposed Cycle-CenterNet also obtains an absolute improvement by 24.6 points. 

Our contributions are summarized as follows:
\begin{itemize}
    \item[-] We build a large-scale dataset in wild complex scenes, which provides a variety of new challenges for table structure parsing with several real image distortions. 
    \item[-] {We present an approach Cycle-CenterNet by exploiting the cycle-pairing module optimization with a novel pairing loss proposed, which enables us to precisely group the discrete cells into the structured tables. }
    \item[-] {In the experiments, our method improves the performance of table structure parsing on the WTW dataset by large margins. It also outperforms the state-of-the-art methods on the ICDAR2019 dataset, and achieves competitive results on the ICDAR2013 dataset.}
\end{itemize}

%------------------------------------------------------------------------
\section{Related Work}
\subsection{Existing Datasets}
% In the problem of table structure parsing or recognition, there have been several important datasets including ICDAR-2013~\cite{gobel2013icdar}, 
% Datasets play important roles on benchmarking 
There are a number of datasets including UNLV~\cite{shahab2010open}, ICDAR-2013~\cite{gobel2013icdar}, SciTSR~\cite{chi2019complicated},  PubTabNet~\cite{zhong2019image} and TableBank \cite{li2019tablebank} etc. that are available in table structure parsing. Prior to the deep table structure parsing approaches emerged, the dataset UNLV~\cite{shahab2010open} and ICDAR-2013~\cite{gobel2013icdar} were designed for benchmarking the table structure recognition systems with a limited number (less than 1,000) of tabular images and annotations. To meet the requirement of designing data-driven learning approaches for table structure parsing, large-scale datasets of PubTabNet~\cite{zhong2019image} and TableBank~\cite{li2019tablebank} are proposed, but the incomplete annotations still hinder their development. Recently, FinTabNet~\cite{zheng2021global} and SciTSR~\cite{chi2019complicated} datasets add the cell coordinates and row-column information to become the most complete and large-scale dataset for table structure parsing task.

Although the scale of table image datasets has been dramatically improved, these datasets are with a specific focus on the document images that are obtained from the digital documents (\eg, PDF documents). For our purpose of parsing table structures in the wild, those datasets cannot be used for training a learning-based approach with an expected generalization ability. Recently, a new dataset ICDAR2019 \cite{gao2019icdar} introduce a more challenging task of parsing table structures from the scanned archival documents instead of the digital documents. However, it only contains 750 images for table structure parsing, which will induce the same problem for training a data-driven table structure parsing model. Besides, the ICDAR2019~\cite{gao2019icdar} dataset is still focused on the document images. 
In contrast to those existing datasets, we contribute a new large-scale table dataset WTW that contains 14,581 complex wired tables in multiple real scenes including photoing, scanning, and web pages. Different from the existing datasets, the images in our proposed dataset usually contain severe practical image distortions including bending, tilting, and occlusion, \etc.

\subsection{Table Structure Recognition and Parsing}
The problem of table structure parsing was previously studied as two sub-problems of table detection and table structure recognition. Kieninger~\etal~\cite{kieninger1998table} presented the first table structure recognition system that estimates the table structures by clustering the detecting the text blocks from tabular images in a heuristic way. Following this work, the rule-based heuristic approaches were proposed~\cite{wang2004table,tengli2004learning} to recognize or parse table structures from the hand-crafted visual cues. Recently, the deep learning-based approaches were proposed to automatically learn informative visual features~\cite{gobel2013icdar,chi2019complicated,zhong2019image,prasad2020cascadetabnet}. 
However, these methods mainly focuses on the well-conditioned document images, where the tables~\cite{schreiber2017deepdesrt,zheng2021global,prasad2020cascadetabnet,li2019tablebank,zheng2021global,raja2020table} are well-aligned to the image axes. As this assumption does not hold on to more challenging tables, some researches tried to get rid of the well-aligned assumption and modeled the table structure parsing problem with graph convolution networks ~\cite{chi2019complicated,qasim2019rethinking,zhong2019image}. However, they are implicitly using the adjacent relationship between the detected cells to construct an informative initial graph and then prune the unexpected edges during training. Different from those approaches, our proposed Cycle-CenterNet gets rid of using the assumptions mentioned above to meet more practical requirements for table structure parsing in the wild.

%------------------------------------------------------------------------
\section{The WTW Dataset}

This section presents the detail of our proposed dataset, {\em Wired Tables in the Wild} (WTW), which has a total of 14581 images in a wide range of real business scenarios and the corresponding full annotation (including cell coordinates and row/column information) of tables.

\subsection{Image Collection and Annotation}

The images in the WTW dataset are mainly collected from the natural images that contain at least one table. As our purpose is to parsing table structures without considering the image source, we additionally add the archival document images and the printed document images. Statically, the portion of images from natural scenes, archival, and printed document images are 50\%, 30\%, and 20\%. After obtaining all the images, we statically found 7 challenging cases.
%, the exemplar images in those cases are shown in Fig.~\ref{fig:2}. 
As summarized in Tab.~\ref{tab_dataset}, our proposed WTW dataset covers all challenging cases with a reasonable proportion of each case. 

% Compared to the most existing datasets~\cite{} that are build on the printed document images, the images in our dataset are collected from Internet in a wide range of real business scenarios. 
% To the best of our knowledge, this dataset is the first and the largest dataset for table structure parsing in natural images. A summary of our dataset is reported in Tab.~\ref{tab_dataset}.

In our dataset, we annotate all the tables presented in each image for their cell coordinates and the row/column information. For the images that have more than one table, their instance information is also annotated. When annotating the cell coordinate, we follow the benchmark of ICDAR2019~\cite{gao2019icdar} to use the inner table lines for localization. To ensure that there is no leakage of sensitive information (names, telephone numbers, etc.), we erased them out. 

% \vspace{-10pt}
\paragraph{Data splitting.} In order to ensure that the training data and test data distributions approximately match, we randomly select approximately 75\% of the original images as the training set, and the rest data samples are used for testing and evaluation. Finally, our WTW dataset has 10970 training samples and 3611 testing ones.
% In each subset, we ensure that the distribution of the challenging cases mentioned in Tab.~\ref{tab_dataset} are similar. 

% \subsection{Tasks}
% WTW dataset can serve the table detection and table structure recognition tasks. Table detection is to locate the table's bounding box, and table structure recognition need to get the table cells coordinates (physical structures) and their row/column information (logical structures). This paper aims to simultaneously detect tables and parsing table structure from WTW dataset. Here, the most challenging problem brought by WTW is how to determine the table id of each predicted cells, and how to group these discrete cells together. We designed a pipeline to solve this problem.

% \subsection{Pipeline and Evaluation Metric}
\begin{table}[t!]
    \centering
    \caption{A statistical summary and comparison between our WTW dataset and the existing datasets for table structure parsing. Our proposed dataset covers all the 7 challenging cases of (1) Inclined tables, (2) Curved tables, (3) Occluded tables or blurred tables (short in \emph{Occ. or Blur}, (4) Extreme aspect ratio tables (short in \emph{Ex. AR}), (5) Overlaid tables, (6) Multi-color tables and (7) Irregular tables in table structure recognition. In the last row, we report the total number of samples for all those datasets.}
    \resizebox{1.0\linewidth}{!}{
    \begin{tabular}{c|cccccc|r}
    \toprule
     \makecell{Challenging\\Cases} &  
     \makecell{\footnotesize Tablebank\\ \cite{li2019tablebank}} & 
     \makecell{\footnotesize UNLV \\ \cite{shahab2010open}} & 
     \makecell{\footnotesize Marmot\\
            \cite{paliwal2019tablenet}} & 
     \makecell{\footnotesize SciTSR \\ \cite{chi2019complicated}} & 
     \makecell{\footnotesize ICDAR-13\\
     \cite{gobel2013icdar}} & 
     \makecell{\footnotesize ICDAR-19\\ \cite{gao2019icdar}} & 
     \makecell{\footnotesize WTW \\ (Ours)}\\\midrule
    Inclined    & - &\checkmark  & - & - & - & \checkmark & {\checkmark} \\
    Curved      & - & - & - & - & - & - &  {\checkmark} \\
    {Occ. or Blur} & - & - & - & - & -&\checkmark & {\checkmark} \\
    {EX. AR}      & - & - & - & - & - & - &  {\checkmark} \\
    {Overlaid}      & - & - & - & - & - & - &  {\checkmark} \\
    Multi Color      &\checkmark & - & - & - & \checkmark & - &  {\checkmark} \\
    Irregular      & \checkmark & - & \checkmark & - & - & \checkmark
     &  {\checkmark} \\\midrule
    \# Samples      & 145,000 & 423 & 1,000 & 15,000 & 156 & 750 &  14,581 \\\bottomrule
    \end{tabular}
    }
 \label{tab_dataset}
\end{table}

\subsection{Baselines and Benchmark Evaluation}
\label{Pipeline and Evaluation Metric}

As our dataset contains a large number of tables that have deformations (\eg, the inclined, curved, and irregular tables), the commonly-used rectangular representation of tables cannot be generalized well to those challenging cases in our dataset. As a result, it would be problematic to directly leverage the state-of-the-art data-driven approaches designed for documents in our dataset. Accordingly, we present a more appropriate way to set up the baseline approaches and provide a comprehensive evaluation protocol to benchmark the new approaches on the WTW dataset. 

% \vspace{-10pt}
\paragraph{Baseline configuration.} Instead of modeling the table structures in natural images as large rectangles, we first represent the tabular cells as small objects since the small objects are more robust to the severe image deformation. Based on this, we formulate the problem of table structure parsing in two steps: (1) using a state-of-the-art object detector for cell detection, and then (2) grouping the detected cells into tables by heuristically calculate the spatial proximity between cells. After obtaining the table structures, a post-processing step is applied to the row/column information\footnote{More detail and pseudo-codes for the heuristic grouping scheme and the post-processing module are described in the supplementary materials.}. To make the baseline configuration more convincing, we use four widely-used object detectors of Faster-RCNN~\cite{Faster-RCNN}, TridenNet~\cite{li2019scale}, Cascade-RCNN~\cite{cai2018cascade} and the anchor-free detector CenterNet~\cite{zhou2019objects} as the cell detector in our baseline.

% First, use the object detection methods \cite{wang2017fast,cai2018cascade,li2019scale,zhou2019objects} as our baseline models to detect the cell bounding boxes, then use the distance-based rules to splice all the cells together, here we will get the complete table and table id if all the cells can't be grouped into a same one. At last, use the Parsing-Processing module to get the row/column information. More detail and pseudo codes for distance-based rules and Parsing-Processing module will be given in the supplementary materials. 

% \vspace{-10pt}
\paragraph{Evaluation protocol for table structure parsing.} A reasonable evaluation protocol is important for quantitatively compare different approaches. We evaluate a given table structure parser in two aspects of (1) \emph{the correctness of physical structure} and (2) \emph{the correctness of logical structure}, described as follow:

% \vspace{-5pt}
\begin{itemize}%\compresslist
    \item[-] \emph{Precision, Recall and F-score for physical structure estimation.}
    
    We evaluate the accuracy of cell detection by calculating the precision, recall, and F1-score for the parsing results with regard to the ground truth in the testing split of the WTW dataset. Different from the general object detection, the table structure parsing requires more accuracy of tabular cells with low tolerance. Therefore, the detected cells whose IOU is below $0.9$ are regarded as false positive detections. 
    
    \item[-] \emph{Precision, Recall, F-score and TEDS~\cite{zhong2019image} for adjacent relationship estimation.}

    For logical structure correctness, we follow the evaluation protocol used in document images by calculating the precision, recall, and F-score for the cell adjacency~\cite{gobel2012methodology} and the tree-edit-distance similarity (TEDS)~\cite{zhong2019image}.
    % use TEDS \cite{zhong2019image} and Cell Adjacency Relation-based table structure evaluation \cite{gobel2012methodology}: Recall (Rec.), Precision (Prec.) and F1-metric op (F1) to evaluate the rebuild structure quality. Additional detail are available in
\end{itemize}
% Since cell detection and the structural estimation determines the accuracy of the parsing results together, we evaluate the precision, recall and F-score in two aspects of cell detection and the adjacent relationship estimation. 

% To evaluate the performance of baseline models, we use different metrics. For physical structure (cell coordinates), we use the object detection evaluation: character-level Recall (Rec.), Precision (Prec.) and F1-metric (F1) to evaluate the methods, where we choose IOU=0.9.  \cite{chi2019complicated,gobel2013icdar,shahab2010open,zhong2019image,gobel2012methodology}
% \vspace{-10pt}
\paragraph{Results of baseline models.}
% \subsection{Pipeline Experiments}\label{Pipeline Experiments}
We train the cell detectors\footnote{The training detail is described in the supplemental material.} in the baseline and then parse the table structures with the above-mentioned post-processing schemes. 
Tab.~\ref{tab_pipeline} shows the evaluation results of all baseline models on the testing split of our WTW dataset. Compared with the anchor-based approaches Faster-RCNN, TridenNet, and Cascade-RCNN, the anchor-free CenterNet obtains the best F-score in the aspect of physical structure accuracy as it does not require fine adjustment of parameters. With more accurate cell detection results, it also achieves the best performance for logical structure correctness.

\begin{figure}
    \centering
    \resizebox{\linewidth}{!}{
    \begin{tabular}{cccc}
    \includegraphics[width=0.24\linewidth]{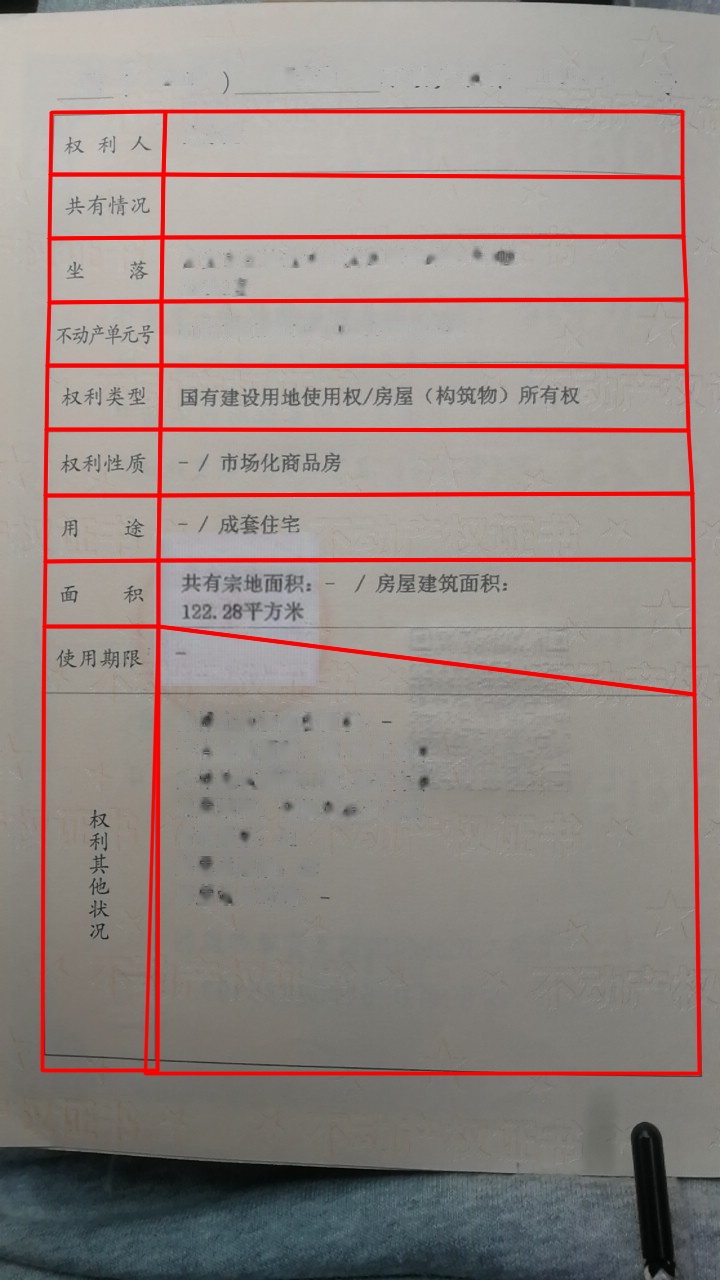} &  \includegraphics[width=0.24\linewidth]{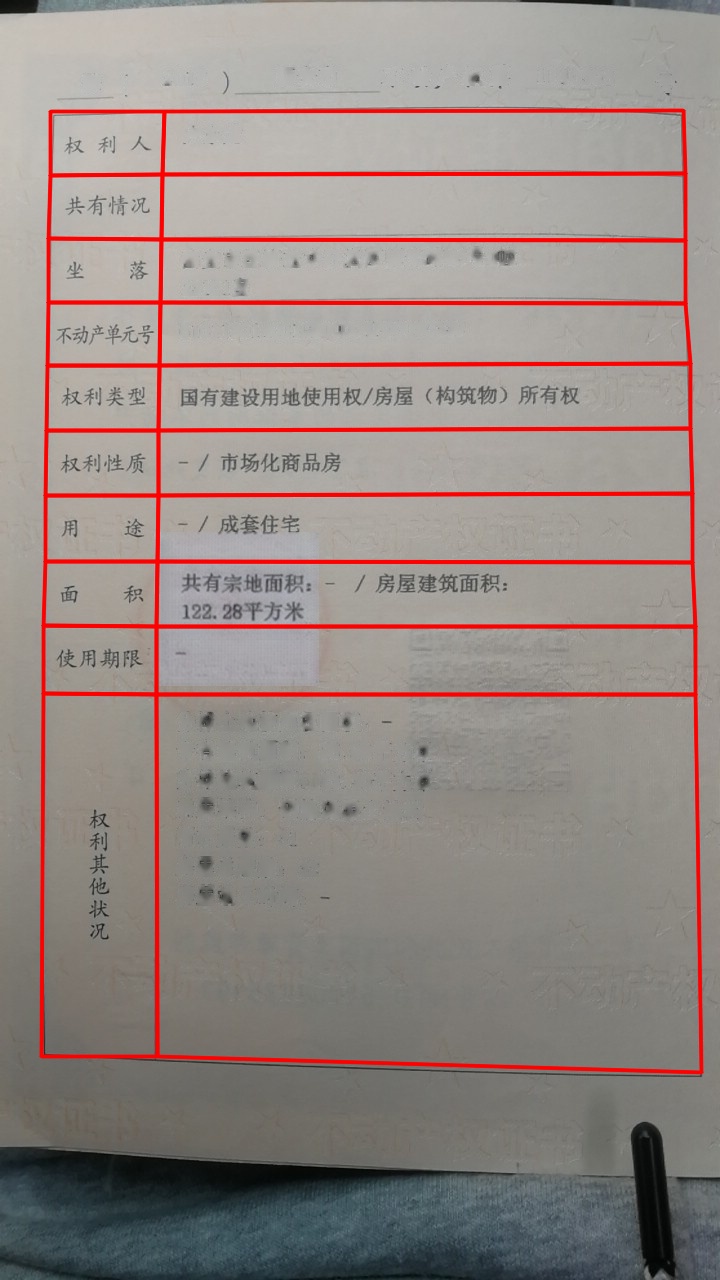} &
    \includegraphics[width=0.24\linewidth]{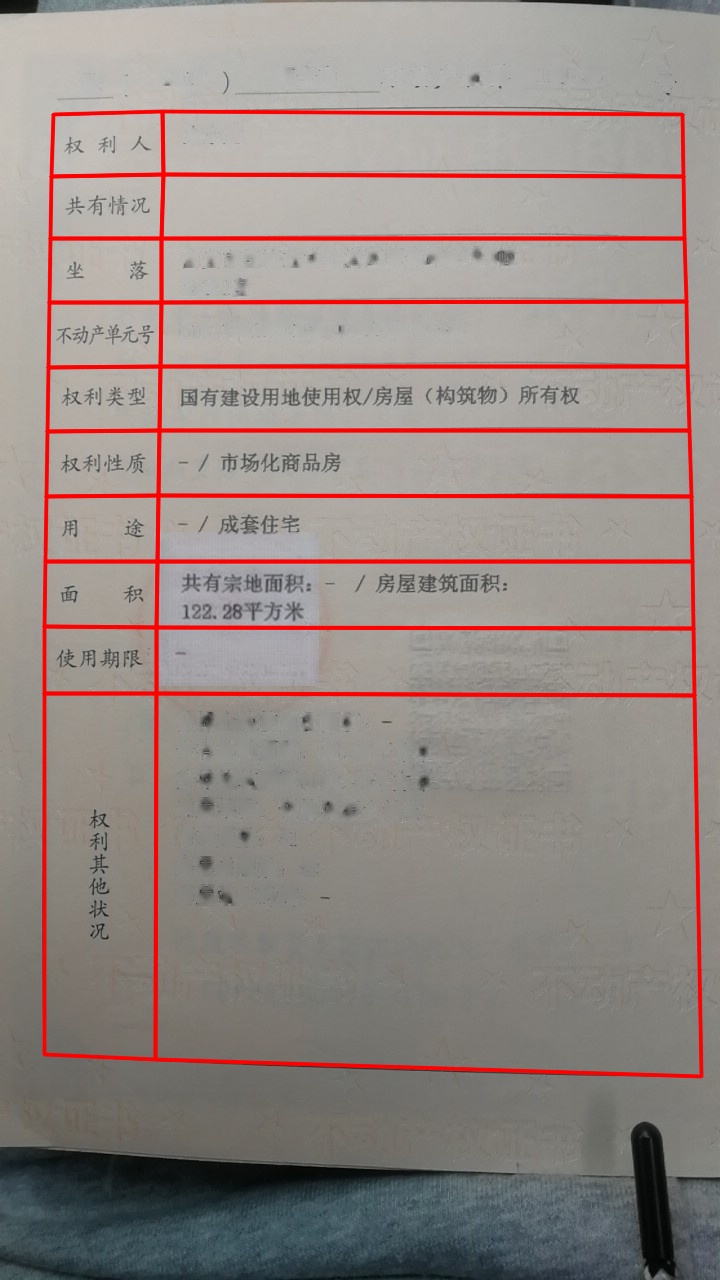} &
    \includegraphics[width=0.24\linewidth]{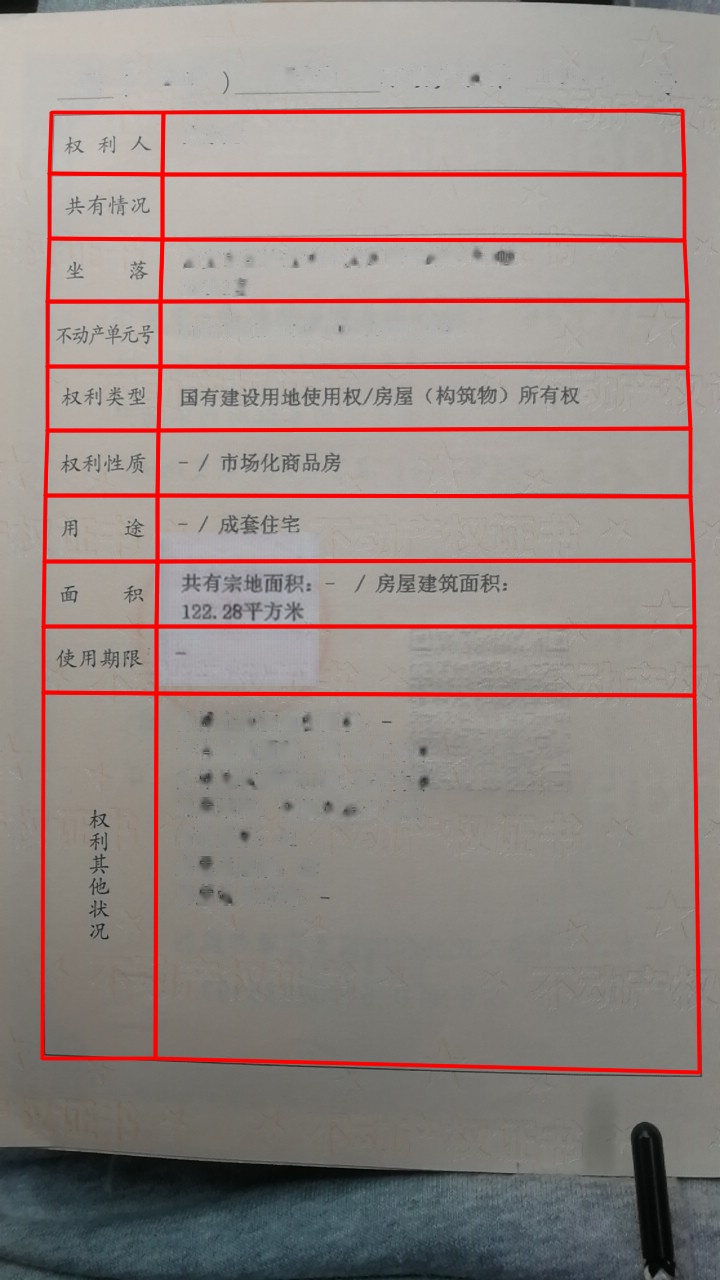} \\
        \includegraphics[width=0.24\linewidth]{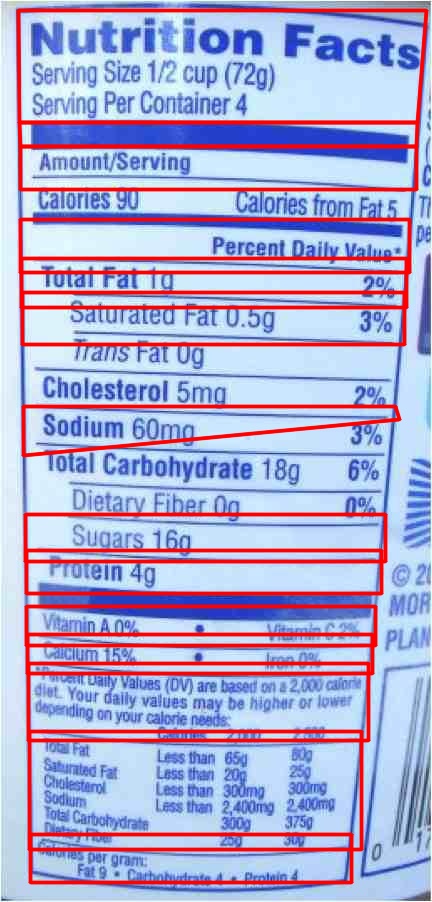} &  \includegraphics[width=0.24\linewidth]{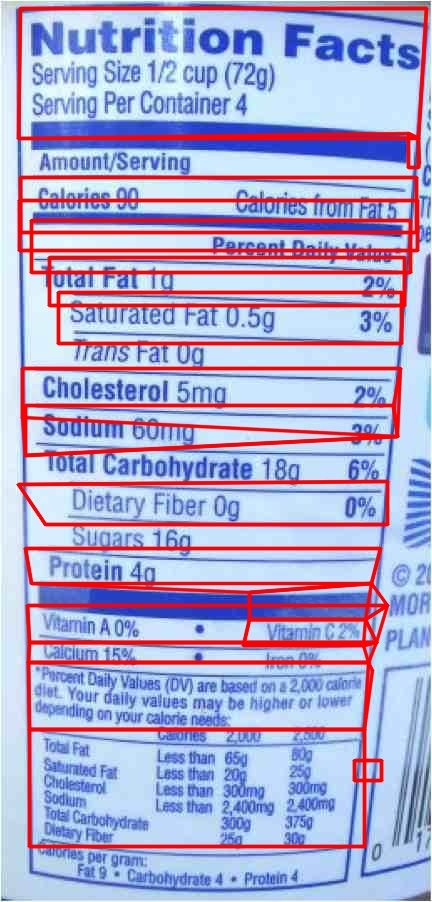} &
    \includegraphics[width=0.24\linewidth]{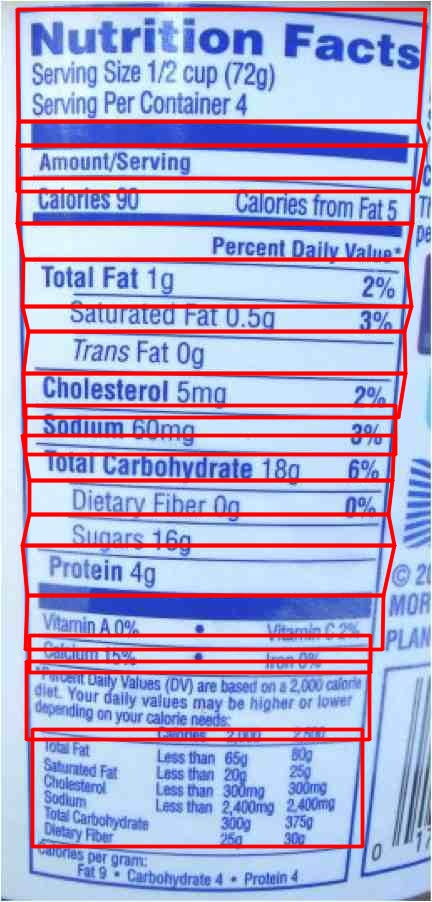} &
    \includegraphics[width=0.24\linewidth]{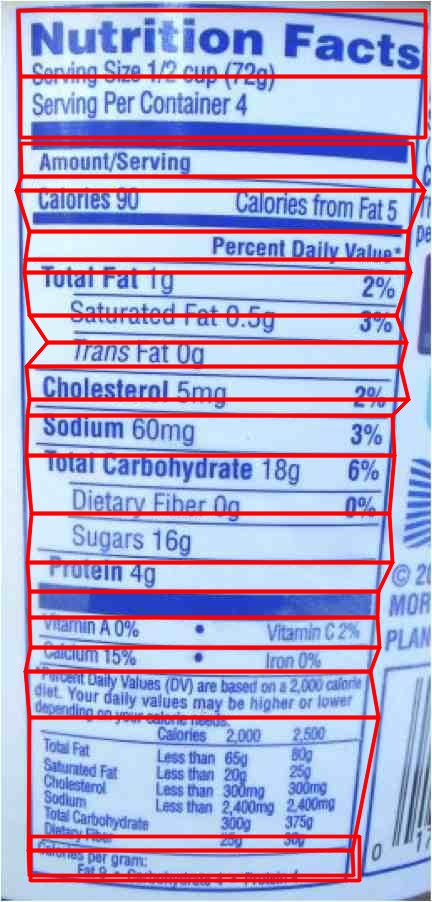} \\
    Faster-RCNN & TridenNet & Cascade-RCNN & CenterNet
    \end{tabular}
    }
    \vspace{1mm}
    \caption{Visualization of the table structure parsing results of the baseline models on the WTW dataset.}
    \label{fig:baseline-vis}
\end{figure}

To further analyze the challenges of table structure parsing in the wild, we visualize the parsing results of two example images taken from different scenes for the baseline models with different object detectors in Fig.~\ref{fig:baseline-vis}. As shown in this figure, the cells can be detected well for both the anchor-based and the anchor-free approaches when the table is approximately aligned with the image domain. By contrast, when leveraging those approaches in the image that have non-rigid deformation, the anchor-based approaches will yield incorrect results. The anchor-free detector, CenterNet, performs better than others while still remaining room for better table parsing accuracy. For more challenging images, it would be of great interest in developing a robust table structure parsing approach.

%\section{Cycle-CenterNet}
%The detail of the proposed Cycle-CenterNet for parsing table in the wild will be discussed in this section. Inspired by Centernet~\cite{zhou2019objects},  we propose the Cycle-CenterNet to learning the belonging information both from the vertexes of each table cell to its center point and splicing relation from the center points of neighbour cells to the common vertex. The pipeline of Cycle-CenterNet is demonstrated in Fig.~\ref{fig:4}. Taking the feature map as input, we predict four branches simultaneously, including a heatmap $Y\in {{R}^{\frac{h}{4}\times \frac{w}{4}\times 2}}$, an offset map $O\in {{R}^{\frac{h}{4}\times \frac{w}{4}\times \text{2}}}$, a center-to-vertex map ${CV_{map}}\in {{R}^{\frac{h}{4}\times \frac{w}{4}\times \text{8}}}$, and a vertex-to-center map ${VC_{map}}\in {{R}^{\frac{h}{4}\times \frac{w}{4}\times \text{8}}}$. The heatmap is the segmentation map of center points and vertexes of table cells. The definition of offset map is the same as Centernet~\cite{zhou2019objects}. With the center-to-vertex information, all the cells of table can be extracted, and the complete table structure could be inferred with the local vertex-to-center splicing information. A Cycle-Pairing module is proposed to model the circular belonging relations, which is jointly optimized by a proposed pairing loss. Besides, a Parsing-Processing is also proposed to get final row/col information. 
\begin{table}[t!]
\small
\begin{center}
\resizebox{1\linewidth}{!}{
\begin{tabular}{|c|c|c|c|c|c|c|c|}
\hline
\multirow{2}*{Model} &
\multicolumn{3}{c|}{Physical Structure} &\multicolumn{3}{c|}{Adjacency Relation} &\multirow{2}*{TEDS}\\
\cline{2-7}
\multicolumn{1}{|c|}{}&Prec.&Rec.&F1&Prec.&Rec.&F1&\multicolumn{1}{c|}{}\\
\hline
Faster-RCNN& 72.1 &61.5 &66.4 & 87.1 & 61.3 & 71.9 &49.5\\
TridenNet& 64.5 &65.5 &65.0 & 85.4 & 71.5 & 77.8 &47.8\\
Cascade-RCNN& 77.4 &65.3 &70.9 & 89.1 & 64.5 & 74.9  &53.2\\\hline
CenterNet&74.2 &72.1 &73.1 & 90.8 & 79.7 & 84.8 &58.7\\
\hline
\end{tabular}
}
\end{center}
\caption{Baseline models on WTW dataset. The physical structure is to measure the accuracy of cell coordinate when IOU=0.9, Adjacency Relation, and TEDS measure the row/col structure information, where the Adjacency Relation is based on the IOU=0.6}
\label{tab_pipeline}
\end{table}

%\section{Cycle-CenterNet}
 \begin{figure*}[t!]
 \centering
 \subfigure{\includegraphics[width=1\linewidth]{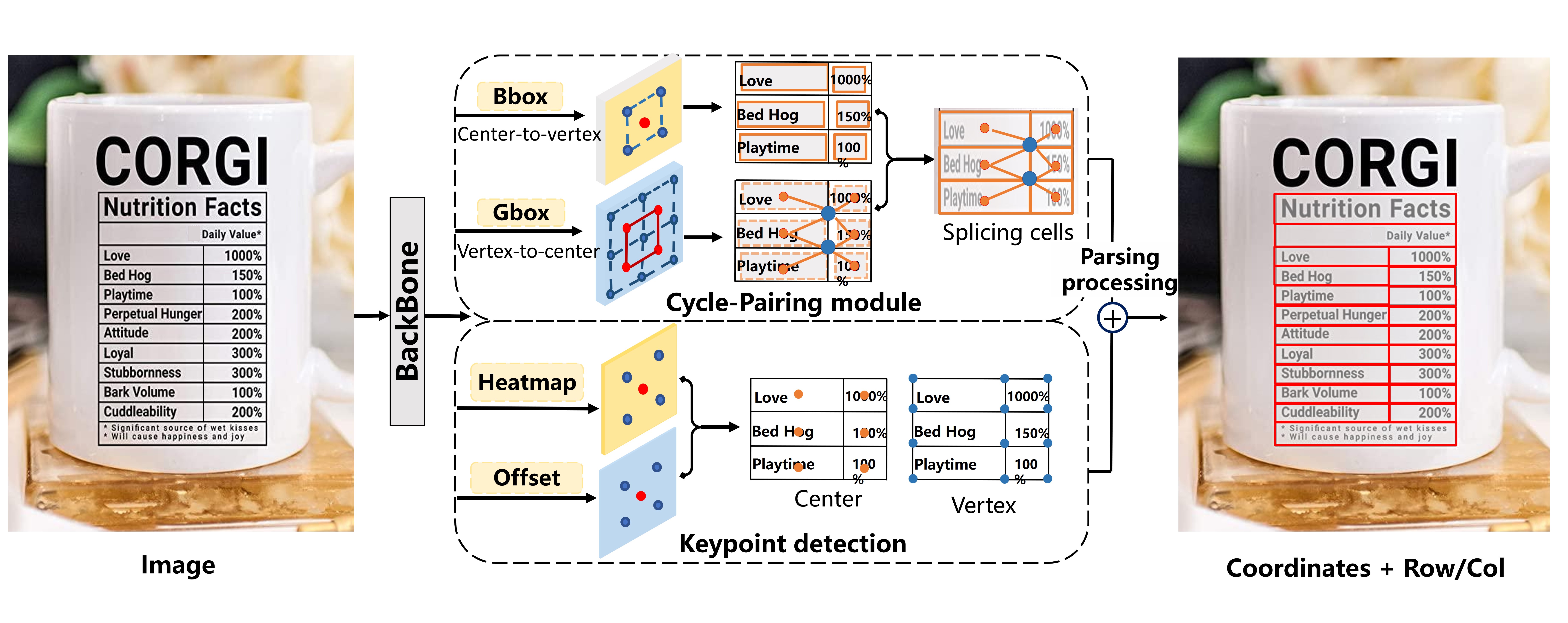}}
 \caption{
 %The pipeline of Cycle-CenterNet. Takeing an image as input, our model produces one heatmap and fed into module behind. Keypoint detection module produces two 2-channel heatmaps, indicate center and vertex. Cycle-Pairing module outputs two 8-channel heatmaps which learns mutual-directed relationship between center point and vertices. According to the relationship, cells grouped and finally output its row and column information by parsing processing. 
 The pipeline of Cycle-CenterNet. Taking an image as input, our model produces one 2-channel keypoint heatmap and one 2-channel offset map. The Cycle-Pairing module outputs two 8-channel heatmaps, which learn the mutual-directed relationship between center point and vertices. According to the relationship, cells are grouped, and finally, the number of row and column information can be recovered by parsing processing. 
 }
 \label{fig:3}
 \end{figure*}

\section{Cycle-CenterNet}
% Inspired by the simplicity of Centernet~\cite{zhou2019objects},  we propose our Cycle-CenterNet for table structure parsing. The network architecture is demonstrated in Fig.~\ref{fig:4}. 
% Taking the feature map as input, we simultaneously predict four branches, including a heatmap $Y\in {{R}^{\frac{h}{4}\times \frac{w}{4}\times 2}}$, an offset map $O\in {{R}^{\frac{h}{4}\times \frac{w}{4}\times \text{2}}}$, a center-to-vertex map ${CV_{map}}\in {{R}^{\frac{h}{4}\times \frac{w}{4}\times \text{8}}}$, and a vertex-to-center map ${VC_{map}}\in {{R}^{\frac{h}{4}\times \frac{w}{4}\times \text{8}}}$. The heatmap is the segmentation map of center points and vertexes of table cells. The definition of offset map is the same as Centernet~\cite{zhou2019objects}. 
% The center-to-vertex map is to learn the belonging information (bounding box) from the center point of each cell to its vertexes, and vertex-to-center map will get splicing relation from the common vertex to its surrounding cell center points. With the center-to-vertex information, all the cells of a table can be extracted, and the complete table structure is inferred with the local vertex-to-center splicing information. 
% A Cycle-Pairing module is proposed by these two branches, which is jointly optimized by a new  pairing loss. Besides, a Parsing-Processing is also proposed to get final row/col information. 

Building on the top of CenterNet, our proposed network adds a Cycle-Pairing module and Pairing loss to learn the common vertex between neighbor cells on the basis of CenterNet~\cite{zhou2019objects}. Through the common vertex, we can splice all the cells together and get a complete table structure. Finally, using the same Parsing-Processing to get row/col information. An illustrative demonstration for our Cycle-CenterNet is shown in Fig.~\ref{fig:3}.

\begin{figure}[t!]
\centering
\subfigure{\includegraphics[width=0.9\linewidth]{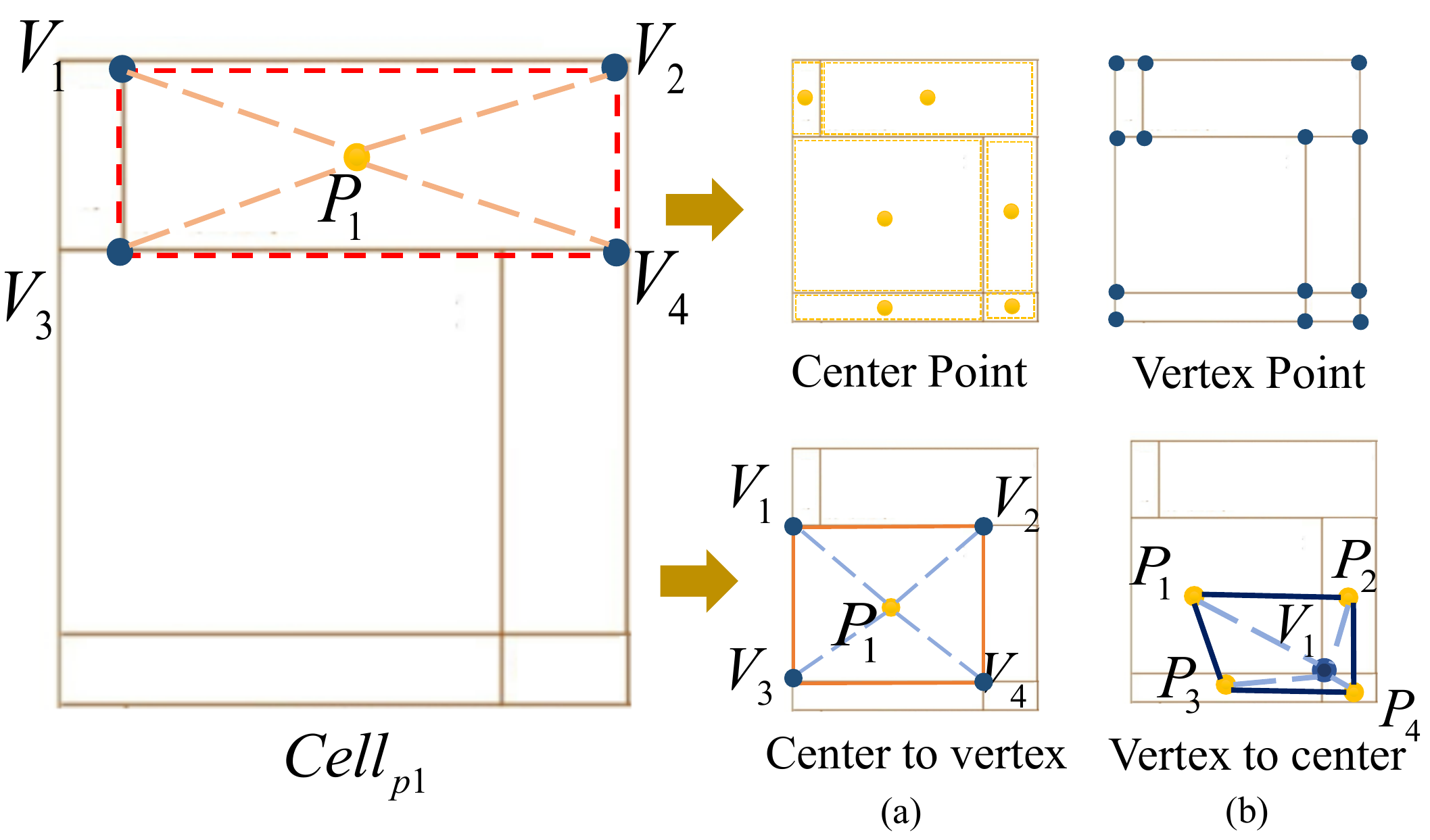}}
\caption{Illustration the process of grouping cells by mutual-directed relationship learned by cycle-pairing module.}
\label{fig:4}
\end{figure}
%-----------------------------------------------------------------
\subsection{Cycle-Pairing Module}
To recognize table structure, we proposed a Cycle-Pairing module to locate the cells and learn the splicing information between cells, which consist of two branches, including \emph{center-to-vertex branch} and \emph{vertex-to-center branch}. As shown in Fig.~\ref{fig:3}, in the center-to-vertex branch, we regress the offset from a center of a table cell to its vertices, and following the post process of Centernet~\cite{zhou2019objects}, the polygonal representation of table cells could be obtained; In the vertex-to-center branch, the offsets between common vertex and the centers of its surrounding cells centers are learned. Finally, the splicing information of tables could be deduced in the Parsing-Processing. %As shown in Fig.~\ref{fig:3}, the details are described as follows:

%through computing the distance from common vertex ${{V}_{1}}$ to surrounding center points ${{P}_{1}},{{P}_{2}},{{P}_{3}},{{P}_{4}}$ we can splice these neighbour cells together, Here are the detail:

\paragraph{Center-to-Vertex branch for cells localization.} Taking the feature map $F$ from the DLA-34~\cite{DLA34} backbone as input, the center-to-vertex branch predict a ${{CV}_{map}}\in {{R}^{\frac{h}{4}\times \frac{w}{4}\times 8}}$. As shown in Fig.~\ref{fig:4} (a), the ${{CV}_{map}}$ indicates the coordinate offset ${\{\Delta{x}, \Delta{y}\}}$ between the center point ${P = \{x_{C}, y_{C}\}}$ and its four vertices ${V = \{x_{V},y_{V}\}}$, denoted by 
\begin{equation}\label{eq5}
\left\{ \begin{split}
  & \Delta {{x}_{{{C}_{ik}}}}={{x}_{{{C}_{i}}}}-{{x}_{{{V}_{ik}}}} \\ 
 & \Delta {{y}_{{{C}_{ik}}}}={{y}_{{{C}_{i}}}}-{{y}_{{{V}_{ik}}}} \\ 
\end{split} \right.{,{i=1:{{N}^{C}},k=1:4}}\text{ },
\end{equation}
where ${{N}^{C}}$ is the number of all the center points of table cells. 
% Through the center point coordinates and coordinate distance, we can get all the four vertices belongs to one cell, it is called cell bounding box (bbox).

% \textbf{Vertex-to-center branch} given the feature map $F$ and send it into Vertex to center branch to get a ${{VC}_{map}}\in {{R}^{\frac{h}{4}\times \frac{w}{4}\times 8}}$, which indicates the offset pair ${\Delta{x}, \Delta{y}}$ channels mean to regress the coordinate distance from common vertex to the surrounding center points:
% \begin{equation}\label{eq6}
% \left\{ \begin{split}
%   & \Delta {{x}_{{{V}_{ik}}}}'={{x}_{{{P}_{i}}}}'-{{x}_{{{V}_{k}}}}' \\ 
%  & \Delta {{y}_{{{V}_{ik}}}}'={{y}_{{{P}_{i}}}}'-{{y}_{{{V}_{k}}}}' \\ 
% \end{split} \right.{{|}_{i=1:{{N}^{C}},k=1:{{N}^{V}}}}\text{ }\
% \end{equation}
% In fact, a common vertex can belong to less than four cells. When the cells sharing the same vertex are less than 4, the coordinate distance is set to 0. Through the center point coordinates and ${{VC}_{map}}$, we can get some center points sharing the same vertex, it is also the splicing information when rebuilding a complete table, which is called group box (gbox).

\paragraph{Vertex-to-Center branch for cells grouping.} Taking the feature map $F$ from backbone DLA-34 as input, the Vertex-to-Center branch predicts a ${{VC}_{map}}\in {{R}^{\frac{h}{4}\times \frac{w}{4}\times 8}}$. As shown in Fig.~\ref{fig:4} (b), the ${{VC}_{map}}$ encodes the coordinate offset ${\{\Delta{x}, \Delta{y}\}}$ between the common vertex ${V = \{x_{V}, y_{V}\}}$ and four center points ${P = \{x_{C},y_{C}\}}$ of the surrounding table cells, denoted by
\begin{equation}%\label{eq6}
\left\{ \begin{split}
  & \Delta {{x}_{{{V}_{ik}}}}={{x}_{{{V}_{i}}}}-{{x}_{{{C}_{ik}}}} \\ 
 & \Delta {{y}_{{{V}_{ik}}}}={{y}_{{{V}_{i}}}}-{{y}_{{{C}_{ik}}}} \\ 
\end{split} \right.{{|}{i=1:{{N}^{V}},k=1:4}}\text{ },
\end{equation}
where ${{N}^{V}}$ denotes the number of all the common vertexes. If the number of cells sharing this vertex $K$ is smaller than 4, the regression value of the remaining positions are set to 0.

\subsection{Pairing Loss for Cycle-Pairing Module} 
% To train the ${{VC}_{map}}$ and ${{CV}_{map}}$ together, we argue that through parsing ${{VC}_{map}}$ and ${{CV}_{map}}$ into a mutual-directed center-vertex pair, we can easily train these two separate maps together. So, we define a pairing loss to represent this training process. Pairing loss is composed of ${{l}_{cv}}$, ${{l}_{vc}}$ and weight $W$. We will introduce how to generate them separately.
Instead of directly applying loss functions on the output maps of Cycle-Pairing Module, we design a pairing loss to supervise the network learning better offsets for both center-to-vertex and vertex-to-center branches by pair-wise compute the loss function for each pair of the center and vertex that belongs to the same cell in the expected table. Denoted by $\mathcal{P}_{cv} = (\Delta x_{cv},\Delta y_{cv},\Delta x_{vc}, \Delta y_{vc})$ for the predicted offset of a pair of center $c$ and $v$, we compute the loss function $L_p$ by
\begin{equation}
    L_{p} = \sum_{c,v} \omega(\mathcal{P}_{cv}) \left(\lambda_{cv} L_{cv} + \lambda_{vc} L_{vc}\right),
\end{equation}
where $L_{cv}$ and $L_{vc}$ are the $\ell_1$ loss between the predicted offsets and the corresponding groundtruth, $\lambda_{cv} = 1.0$ and $\lambda_{vc} = 0.5$ are the hyperparameters to tune the importance between those loss items, $\omega(\mathcal{P}_{cv})$ dynamically weighing the overall loss according to the regression quality.

\begin{figure}[t!]
    \subfigtopskip=2pt %设置子图与上面正文或别的内容的距离
% 	\subfigbottomskip=2pt %设置第二行子图与第一行子图的距离，即下面的头与上面的脚的距离
	\subfigcapskip=-6pt %设置子图与子标题之间的距离

    % \begin{minipage}{\linewidth}
    \centering
    \subfigure[$\mathcal{D}_{cv}=0$]{
    \includegraphics[width=0.23\linewidth]{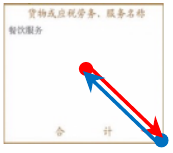}}
    \subfigure[$\mathcal{D}_{cv}\in(0,1)$]{
    \includegraphics[width=0.23\linewidth]{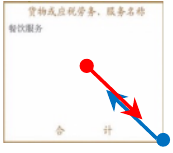}}
    \subfigure[$\mathcal{D}_{cv}=1$]{
    \includegraphics[width=0.23\linewidth]{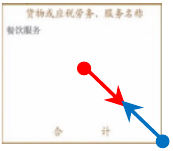}}
    \subfigure[$\mathcal{D}_{cv}>1$]{
    \includegraphics[width=0.23\linewidth]{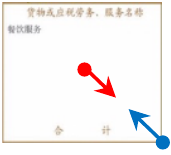}}
    \caption{Illustration of the traditional cases for center-vertex pairs during training.}
    \label{fig:exp-illustrate} 
    % \end{minipage}
    %\vspace{-4mm}
    % }
\end{figure}

\paragraph{Dynamic weighing function $\omega(\mathcal{P}_{cv})$.}
The cycle-pairing module represents the pairwise pointing relationship between vertex and center point. In fact, it is not necessary to regress cell bounding box and common vertex group box such accurately, as long as the prediction of center and vertex from one center-vertex pair are intersected. So we use $\omega(\mathcal{P}_{cv})$ to weight losses ${{l}_{cv}}$ and ${{l}_{vc}}$ for center-vertex pairs:
\begin{equation}\label{eq6}
\omega(\mathcal{P}_{cv}) = 1-{\exp\left(-\pi \mathcal{D}_{cv}\right)},
\end{equation}
where $\mathcal{D}_{cv}$ is the pair distance defined as
% where ${{pa}_{i}}$ is a center-vertex pairs in $P{{A}^{*}}$. We define the Pair Distance $PD({{pa}_{i}})$ to metric the value of $W$:
\begin{equation}\label{eq7}
\mathcal{D}_{cv}=\min (\frac{\left| {{x}_{c{{v}_{i}}}}-{{x}_{c{{v}_{i}}}}^{*} \right|+\left| {{x}_{v{{c}_{i}}}}-{{x}_{v{{c}_{i}}}^{*}} \right|}{\left| {{x}_{c{{v}_{i}}}}^{*} \right|},1)
\end{equation}
where ${{x}_{cv}}$ is a regression value from center to vertex, while ${{x}_{vc}}$ is vertex to the center. Therefore, $\mathcal{D}_{cv}$ defines the regression error score for each center-vertex pair. As shown in Fig.~\ref{fig:exp-illustrate}, if $\mathcal{D}_{cv}=0$, means that the vertex and the center point to each other strictly without any errors. If $0<\mathcal{D}_{cv}<1$, means that although the vertex and center couldn't strictly point to each other, but pointing into each other's bounding box. If $\mathcal{D}_{cv}\ge 1$, means that there is no intersection between the center and vertex, which is the main sample that needs to focus on learning.

%-------------------------------------------------------------------------
%-------------------------------------------------------------------------

% Therefore, we use the regression error score $PD({{pa}})$ to define the weight $W({pa})$ of the loss function, and a truncated Gaussian distribution is used to simulate $W({pa})$ as shown in Eq.~\ref{eq6}. $W({pa})$ is positively correlated with $PD({{pa}})$, which makes model pay attention to learning hard samples.

% In summary, Pairing loss can be described as:
% \begin{equation}\label{eq7}
% {{L}_{p}}=\sum{w({{pa}_{i}})({{\lambda }_{cv}}{{l}_{c{{v}_{i}}}}+{{\lambda }_{vc}}{{l}_{v{{c}_{i}}}})}
% \end{equation}
% where we set ${{\lambda }_{cv}}=1$ and ${{\lambda }_{vc}}=0.5$.

The loss ${{L}_{k}}$ of the keypoint branch and the loss ${{L}_{off}}$ of the offset branch are consistent with CenterNet~\cite{zhou2019objects}. The overall training loss is:
\begin{equation}\label{eq8}
{{L}_{det}}={L}_{k} + {{\lambda }_{off}}{{L}_{off}} + {L}_{p}
\end{equation}

\subsection{Parsing-Processing Module} 
As the last step, we propose a Parsing-Processing module to recover the complete table structure information, including table id, start row/column, and end row/column. 
% todo: fix-me
First, split every cell into 4 bounding edges, then merge the up edges and down edges to horizontal lines and merge left edges and right edges to vertical lines according to cell connectivity. Next, sort the horizontal lines, vertical lines and index them from 0. Finally, rank cells by line index and outputs row/column information. The pseudo code is given in the supplementary materials.

%\subsection{Network Architecture}
% As shown in Fig.~\ref{fig:3}, a wired table image will be sent into backbone DLA-34 to get a feature map $F\in {{R}^{\frac{h}{4}\times \frac{w}{4}\times c}}$, then we use Keypoint detection module and Cycle-Pairing module to detect the cells and learn splicing information. Keypoint detection module is a common detection module following Centernet \cite{zhou2019objects}, it will get a heatmap $Y\in {{R}^{\frac{h}{4}\times \frac{w}{4}\times 2}}$ and an offset $O\in {{R}^{\frac{h}{4}\times \frac{w}{4}\times \text{2}}}$ to locate each cell center point and vertex. It uses the same loss function: keypoint regression and offset loss ${{L}_{\det }}$ to monitor its accuracy, more detail can be found in \cite{zhou2019objects}. And the Cycle-Pairing module with Pairing loss will get the splicing information ${{VC}_{map}}$ and ${{CV}_{map}}$ of all the cells. From the output of Cycle Centernet, we can get a complex table structure which has cells' coordinates information. Finally, we use Parsing-Processing module to get the row and column number of each cell.
 
\subsection{Training Detail}
%We train the models on WTW dataset using 10970 images, testing with another 3611 images and divided into 6 classes - Inclined, Curved, Occluded and blurred, Extreme aspect ratio, Overlaid and Muti color and cell. When training Cycle Centernet, we keep the same parameters with Centernet\cite{zhou2019objects} on coco. To train the model more efficiently, we resize the long side of training image to 1024, and scale the short side equally. Finally, we use 8 GeForce GTX 1080 with 12gb memory for our experiments and a batch-size of 32, set the initial learning rate with \num{1.25E-3} with total 140 epochs, and when reaching the 90th and 120th epoch, changing the learning rate to \num{1.25E-4} and \num{1.25E-5}.

In the training process of the Cycle-Centernet, we use the pre-trained weight on COCO, and resize the max side of the training image to 1024 with scaling the short side equally. The initial learning rate is set to \num{1.25E-3}, and decayed to \num{1.25E-4} and \num{1.25E-5} in the 90th and 120th epoch respectively. The model is trained with a total of 150 epochs. All the experiments are performed on a workstation with 8 NVIDIA GTX 1080Ti GPUs. During the training, we set the batch size to 32 per GPU in parallel. 
%-------------------------------------------------------------------------
\begin{table*}
\small
\begin{center}
\setlength{\tabcolsep}{2.87mm}{
\begin{tabular}{|c|c|c|c|c|c|c|c|c|c|}
\hline
\multirow{2}*{Category}&\multirow{2}*{Model} & \multirow{2}*{Backbone} &
\multicolumn{3}{c|}{Physical coordinates} &\multicolumn{3}{c|}{Adjacency Relation} &\multirow{2}*{TEDS}\\
\cline{4-9}
\multicolumn{1}{|c|}{}&\multicolumn{1}{c|}{}&\multicolumn{1}{c|}{}&Prec.&Rec.&F1&Prec.&Rec.&F1&\multicolumn{1}{c|}{}\\
\hline
Base Model&CenterNet& DLA-34  & 74.2 &72.1 &73.1 & 90.8 & 79.7 & 84.8 &58.7\\
\hline\hline
\multirow{2}*{\makecell{Table structure\\ models}}&Split+Heuristic &-&3.2&3.6&3.4&25.7&29.9&27.6&26.0\\

\multirow{2}*{}&CascadeTabNet
&CascadeNet &-&-&-&16.4&3.6&5.9&11.4\\

\hline\hline
\multirow{3}*{Ours}&CenterNet(Polygon box)& DLA-34 & 75.1 & 75.7 &75.4 & 93.0 & 89.2& 91.1&70.1\\

\multirow{3}*{}&Cycle-CenterNet& DLA-34 & \textbf{78.2}& 78.2 &78.2 & 93.2 &91.4 &92.2  &74.3 \\

\multirow{3}*{}&Cycle-Centernet+PairLoss& DLA-34 & 78.0 &\textbf{78.5} &\textbf{78.3} & \textbf{93.3} & \textbf{91.5} & \textbf{92.4} &\textbf{83.3} \\
\hline
\end{tabular}}
\end{center}
\caption{Results on WTW dataset. The evaluation metrics are same with Tab.~\ref{tab_pipeline}}
\label{tab2}
\end{table*}

\begin{table*}
\small
\centering
%\begin{threeparttable}
\setlength{\tabcolsep}{2.1mm}{
\begin{tabular}{|c|c|c|c|c|c|c|c|c|c|c|c|c|c|c|}
\hline
\multirow{2}*{Method}& \multicolumn{2}{c|}{Simple}& \multicolumn{2}{c|}{Inclined}& \multicolumn{2}{c|}{Curved}& \multicolumn{2}{c|}{\makecell{Occluded \\and blurred}} & \multicolumn{2}{c|}{\makecell{Extreme \\aspect ratio}} & \multicolumn{2}{c|}{Overlaid} & \multicolumn{2}{c|}{\makecell{Muti \\color and grid}}\\
\cline{2-15}
\multicolumn{1}{|c|}{}&F1&TEDS&F1&TEDS&F1&TEDS&F1&TEDS&F1&TEDS&F1&TEDS&F1&TEDS\\
\hline
Ours&99.3&94.2 &97.7 &90.6  &76.1 &70 &77.4 &53.3  &91.9 &77.4 &84.1 &51.2  &93.7 &66.7\\
\hline
\end{tabular}
}
 %\begin{tablenotes}
 %       \footnotesize
 %       \item[*] this is the abbreviation of 2D-TEDS  %此处加入注释*信息
 %      \item[**] this is the abbreviation of TEDS %此处加入注释**信息
 %     \end{tablenotes}
 %     \end{threeparttable}
\caption{Results on different categories.}
\label{tab3}
\end{table*}

\section{Experiments}

We perform extensive experiments on the proposed WTW dataset to verify the effectiveness of Cycle-CenterNet. Although we mainly focus on wired tables in wild scenes, we additionally do experiments on the widely used benchmarks of ICDAR2013 and ICDAR2019 to demonstrate that: 1) the WTW dataset covers a wide range of tabular images for practical applications and 2) our proposed Cycle-CenterNet is able to recognize wireless tables. 

\begin{table}
\small
\begin{center}
\resizebox{0.95\linewidth}{!}{
\begin{tabular}{|c|c|c|c|c|}
\hline
\multirow{2}*{Model} &\multirow{2}*{\makecell{Training\\Datasets}}&\multicolumn{3}{c|}{IOU=0.6}\\
\cline{3-5}
\multicolumn{1}{|c}{}&\multicolumn{1}{|c|}{}&Prec.&Rec.&F1\\
\hline
DeepDeSRT \cite{schreiber2017deepdesrt} &SciTSR &63.1 & 61.9 & 62.5  \\

Split+Heuristic \cite{tensmeyer2019deep} & Private &93.8 & 92.2 & 93.0 \\
TableNet \cite{paliwal2019tablenet}& Marmot Extended &92.2 &89.9  & 91.0 \\
Tabstruct-Net \cite{raja2020table}& SciTSR &91.5 & 89.7 & 90.6  \\
\hline\hline
Ours & WTW+ICDAR19&95.5 & 88.3 & 91.7 \\
Ours* & WTW&97.5 & 98.4 & 98.0 \\

\hline
\end{tabular}
}
\end{center}
\caption{Comparison for cell adjacency relation on ICDAR-2013 dataset. Here "*" denotes for the result on wired tables only.}
\label{tab_ICDAR-2013}
\end{table}

\begin{table}
\small
\begin{center}
\resizebox{0.95\linewidth}{!}{
\begin{tabular}{|c|c|c|c|c|c|c|}
\hline
\multirow{2}*{Model} &\multirow{2}*{\makecell{Training\\Datasets}}&\multicolumn{4}{c|}{IOU}&\multirow{2}*{\makecell{WAvg.}}\\
\cline{3-6}
\multicolumn{1}{|c|}{}&\multicolumn{1}{c|}{}&0.6&0.7&0.8&0.9&\multicolumn{1}{c|}{}\\
\hline
NLPR-PAL & -&36.5 &30.5 & 19.5 & 3.5 &20.6 \\
CascadeTab \cite{prasad2020cascadetabnet}&\makecell{Marmot  etc.}&43.8 &35.4 & 19 &3.6 &23.2 \\
GTE \cite{zheng2021global}& \makecell{FinTabNet}&38.5 & - & - & - &24.8 \\
TabStruct-Net \cite{raja2020table}&\makecell{SciTSR} &80.4 &- & - &- &- \\

\hline\hline
Ours & WTW&\textbf{80.8} & \textbf{51.1} & \textbf{31.9} &\textbf{11.2} &\textbf{40.0} \\

\hline
\end{tabular}
}
\end{center}
\caption{Comparison with participants of ICDAR 19 Track B2 (Modern) F1-scores \cite{gao2019icdar}, here all the methods finetune on the ICDAR-2019 dataset, we just list their initial training datasets.}
\label{tab_ICDAR2019}
\end{table}

\subsection{Evaluation on WTW}

To evaluate the performance of Cycle-Centernet on the WTW dataset, we compare it with some latest table recognition methods including Split+Heuristic \cite{tensmeyer2019deep} and CascadeTabNet \cite{prasad2020cascadetabnet}. For a fair comparison, those methods are retrained on the WTW dataset with the author-provided hyperparameter settings. 
As discussed in Sec.~\ref{Pipeline and Evaluation Metric}, we evaluate the correctness of the parsed tables in both aspects of physical structures and logical structures.  For the physical structure (cell coordinates), the precision, recall and, F1 score are used under a strict IoU threshold of 0.9. For the logical structure (row/column information), TEDS \cite{zhong2019image} and cell adjacent relationship (IOU=0.6) \cite{gobel2012methodology}  are used as evaluation metrics.

% \vspace{-10pt}
\paragraph{Quantitative comparison.} Tab.~\ref{tab2} shows the performance of all the models on the challenging WTW dataset. It is shown that the state-of-the-art table structure parsing approaches CascadeTabNet~\cite{prasad2020cascadetabnet} and Split+Heuristic~\cite{tensmeyer2019deep} for well-conditioned tabular images obtain poor performance on the challenging WTW dataset that has many wild images. By contrast, benefiting from the flexible design of our method, we significantly improve the performance of table structure parsing by large margins.  Our proposed Cycle-CenterNet obtains the best performance by using the proposed pairing loss function.

% Results for Split+Heurist are very intuitive. Split+Heurist is not suitable for wild dataset like WTW, since it is based on the strong assumption that the tables are horizontally (or vertically) aligned. If tables are not completely horizontal, the table lines predicted by Split+Heurist will has a certain offset from the ground truth. CascadeTab-Net also has a poor performance on WTW. Although this result matches our expectations, we still search for the specific reasons. We find that the CascadeTab-Net just uses CascadeNet for table detection on wired tables, as for the structure recognition, it is launched by Opencv and post-processing. Our wild tables require the model to be robust enough, but the post-processing in CascadeTab-Net sets many  hyperparameters, which possibly results in severely overfitting. 

% Limited by the current horizontal and vertical datasets, the existing algorithm performs poorly on the wild tables, which implies the challenging of WTW. Meanwhile, our Cycle-CenterNet obtains a good performance, reaches a 92.4$\%$ f1 score on adjacent relationship, and 83.3$\%$ on TEDS. It is a strong baseline for wired tables in the wild.
\begin{figure*}
\centering
\subfigure{\includegraphics[width=1.0\linewidth]{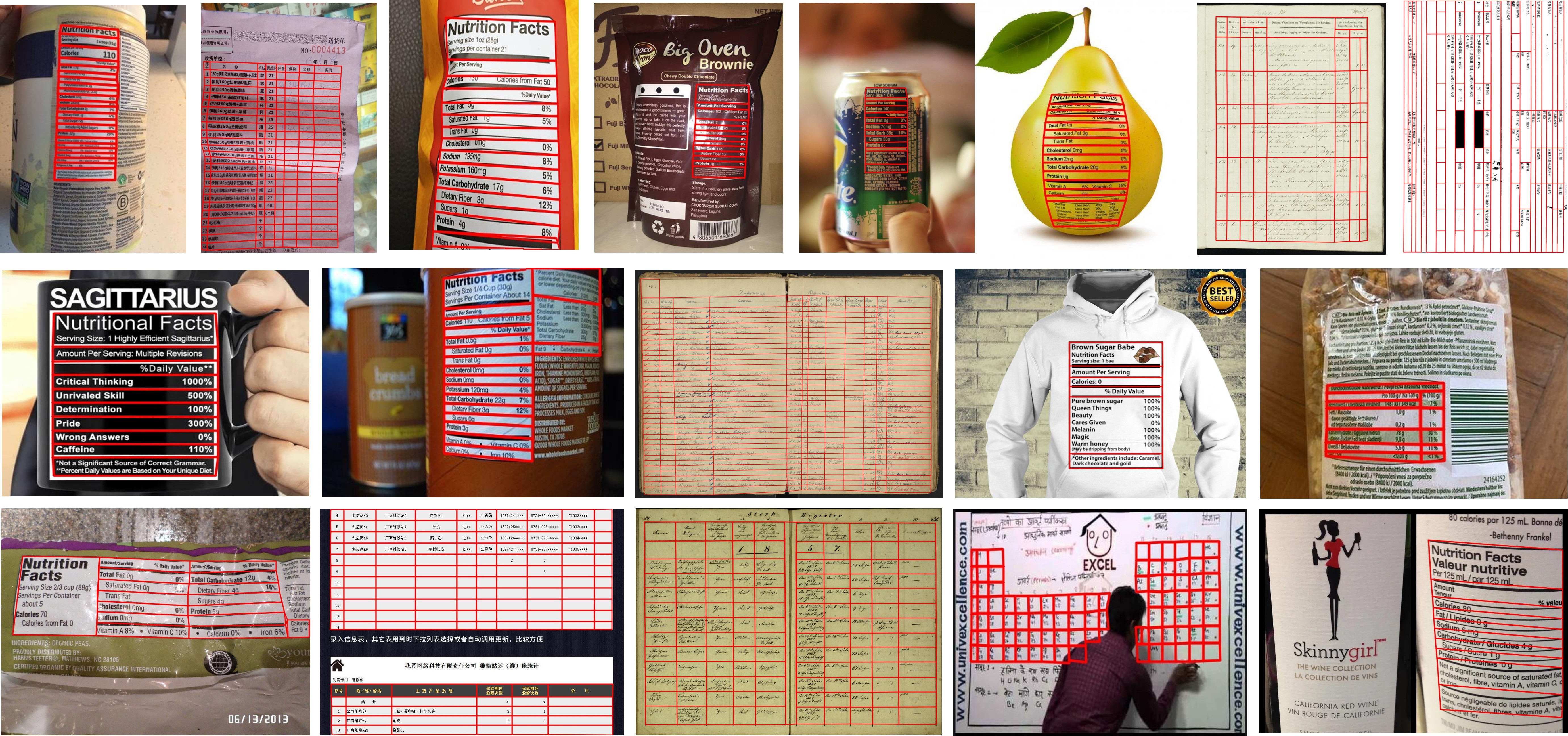}}
\caption{Qualitative results of Cycle-Centernet on different datasets }
\label{fig:5}
\end{figure*}

\vspace{-2mm}
\paragraph{Ablation study.} %We make incremental improvements on CenterNet and obtain the final Cycle-CenterNet, so a series of ablation experiments are made to verify its effectiveness. 
In Sec.~\ref{Pipeline and Evaluation Metric}, we argue that the accuracy of cell regression has a great influence on table structure recognition, so we replace horizontal rectangle regression with arbitrary quadrilateral regression in CenterNet. Although the cell detection is only increased by 2.3$\%$, we get 6.3$\%$ and 11.4$\%$ improvement on adjacency relation and TEDS. The cycle-pairing module can simultaneously detect and group tabular cells into structured digital tables, which makes the Cycle-CenterNet get 2.8$\%$ improvement on cell detection, 1.1$\%$ improvement on adjacency relation, and 4.2$\%$ improvement on TEDS. With the collaborative optimization of Pairing loss on center-vertice pair, the addition of Pairing loss makes Cycle-CenterNet significantly increase on the dedicated table structure evaluation matrix of TEDS by 9$\%$.

Compared with our base model CenterNet, the Cycle-CenterNet has a 7.6$\%$ improvement on adjacency relation and 24.6$\%$ improvement on TEDS. Since the models in other papers are not designed for wild tables, our Cycle-CenterNet shows obvious advantages in all evaluation metrics.

%\begin{figure}
%\centering
%\subfigure{\includegraphics[width=8.5cm]{figs/fig7.png}}
%\caption{Bad samples on the deformed, occluded and blurred tables.}
%\label{fig:7}
%\end{figure}
\vspace{-2mm}
\paragraph{Sub-category experiments.}
To verify the complexity of our WTW dataset, we analyze the model results on different types of tables separately. Results are shown in Tab.~\ref{tab3}. Cycle-CenterNet achieves 99.3$\%$ in adjacency relation and 94.2$\%$ in TEDS on the simple subset of WTW, 97.7$\%$ in adjacency relation and, 90.6$\%$ in TEDS on the inclined subset. It means the Cycle-CenterNet can get a good result for ordinary tables. Relatively good results (91.9$\%$ at adjacency relation and 77.4$\%$ at TEDS) are obtained for tables with extreme aspect ratios for the vertex-to-center branch which can pull the bounding box to the vertex. 
Although our Cycle-Centernet has reached the state of the art, it still needs to be improved on some styles like curved, overlaid, etc. Our model just gets 53.3$\%$ at TEDS on occlusion and blur subset. Besides, multiple table superposition also brings great difficulties to the table recognition task. A table with especially dense cells, combined with any slightly complex feature makes the task drastically difficult. %For example, when there exist dense cells with dense text, then the table line will be covered by text, which results in a particularly poor result. 
We will continue to seek solutions to these problems and look forward to more researchers joining in.

\subsection{Evaluation on other Datasets}
To evaluate the robustness and versatility of Cycle-CenterNet, we test our model on two mostly used datasets in table structure recognition, ICDAR-2013 \cite{gobel2013icdar} and ICDAR-2019 \cite{gao2019icdar}. These two datasets consisting of both wired and wireless tables. For lacking wireless table in WTW, we finetune the Cycle-CenterNet on the ICDAR2019 training set and test on ICDAR 2019 Track B2 dataset. Since ICDAR2013 only has test set, we selected some wireless tables from ICDAR-2019 to finetune our model and test on ICDAR2013. Tab.~ \ref{tab_ICDAR-2013} and \ref{tab_ICDAR2019} show the results on datasets ICDAR-2013 \cite{gobel2013icdar} and ICDAR-2019 \cite{gao2019icdar}. Here the evaluation metric only includes the cell adjacency relation \cite{gobel2012methodology}: Recall (Rec.), Precision (Prec.), F1-metric op (F1). 

Although Cycle-CenterNet is mainly designed for wired tables and trained on WTW, it still obtains 91.7$\%$ fscore on ICDAR2013 with wireless tables, and is only 1.3$\%$ lower than the first-ranked model of Split+Heuristic. This shows our model can also cover wireless tables. Similarly, our WTW mainly focuses on wild wired tables, but Cycle-CenterNet trained only on WTW and tested on ICDAR2013 wired tables can still achieve 98.0$\%$ fscore, indicating that WTW can also cover simple wired tables. 

For ICDAR-2019 \cite{gao2019icdar}, we keep the same evaluation metric: ICDAR 19 Track B2 F1-scores \cite{gao2019icdar}, which is based on the adjacency relation evaluation \cite{gobel2012methodology}. Precision, Recall, and F1 scores are calculated with IoU thresholds 0.6, 0.7, 0.8, and 0.9 respectively. The Weighted-Average F1 (WAvg.) is calculated by assigning a weight to each F1 value of the corresponding IoU threshold. As shown in Tab.~\ref{tab_ICDAR2019}, compared with the highest result of the Weighted-Average F1 reported so far, Cycle-CenterNet has improved 15.2$\%$ and achieve the start of the art.

\section{Conclusions and Future Work}

In this paper, we tackle the problem of table structure parsing in the wild by proposing a new WTW dataset and a deep table structure parser, Cycle-CenterNet. On one hand, the proposed WTW dataset contains about 14k real-scene images that are taken in wild imaging conditions, which pushes the boundary of table structure parsing from the digital document images to the real-scene images. On another hand, we propose a new approach for wild-scene table structure recognition, called Cycle-CenterNet, which addressed the major weaknesses of the existing approaches including imprecise geometry prediction of instances with extremely physical distortion and defectiveness in extracting logical structures of misaligned tables. 
The comprehensive experiments demonstrated that the proposed approach resolves the mentioned issues in a principled way and achieves a new state-of-art for table structure parsing. We hope our proposed WTW dataset can further improve future research on table recognition.
% We also introduce a wild-scene wired table dataset (WTW) to show the widely application of tables and demonstrate the effectiveness of the proposed method. 
% In the future, we are going to involve the wireless tables into our WTW dataset, and we are ready to extend our line-aware approach to a more general version.

% \vspace{-4mm}
\section*{Acknowledgement}%\vspace{-2mm}
{
This work was supported by the National Natural Science Foundation of China under Grant 61922065, Grant 61771350, Grant 41820104006, and National Post-Doctoral Program for Innovative Talents under Grant BX20200248. This work was also supported by Alibaba Group through Alibaba Innovative Research (AIR) program.
The numerical calculations in this paper have been done on the Supercomputing System in the Supercomputing Center of Wuhan University. The views presented in this paper are those of the authors and should not be interpreted as representing any funding agencies.
}
%exist in our work: First of all, our work only restores the structure of the wired form, and cannot identify the structure of the wireless form. Secondly, from a data perspective, despite the limitations of the current algorithm capabilities, complex wired and wireless tables should be split. However, in the future, we can still consider integrating wireless table data, so that more people can explore the method of implementing complex wired tables and wireless tables at the same time, and promote the development of table structure recognition. From the perspective of the model, our current algorithm uses a Parsing-Processing to obtain cell row and column information, in the future, we can explore end-to-end implementation methods, such as directly returning the cell row number and column number, so as to streamline the structure recognition process. In addition, the idea that the vertices and centers of the cycle-centernet point to each other can be used to learn the relationship between two objects at a long distance in space.
%\vfill
%\newpage
%-------------------------------------------------------------------------

%-------------------------------------------------------------------------

%-------------------------------------------------------------------------

{\small
\bibliographystyle{ieee_fullname}
\bibliography{egbib}
}

\end{document}